\documentclass[runningheads]{llncs}

\usepackage{eccv}

\usepackage{eccvabbrv}
\usepackage{amsmath}
\usepackage{multirow}
\usepackage{algorithm}
\usepackage{algorithmic}
\usepackage{amsmath}

\usepackage{graphicx}
\usepackage{booktabs}

\usepackage[accsupp]{axessibility}  

\usepackage[square,numbers]{natbib}
\usepackage{hyperref}
\hypersetup{
    breaklinks=true,colorlinks=true,linkcolor={red!80!black},urlcolor={darkgray!80!black},citecolor={green!80!black}
}
\usepackage{orcidlink}
\usepackage{algorithm}
\usepackage{algorithmic}
\usepackage{subcaption}
\usepackage{marvosym}

\def\rvx{{\mathbf{x}}}
\def\rvX{{\mathbf{X}}}
\def\rvz{{\mathbf{z}}}
\def\rvw{{\mathbf{w}}}
\def\rvW{{\mathbf{W}}}

\def\N{{\mathcal{N}}}
\def\E{{\mathbb{E}}}
\def\KL{{\mathrm{KL}}}
\def\p{p_{\rm data}}

\begin{document}

\title{Learning Multimodal Latent Generative Models with Energy-Based Prior}


\author{Shiyu Yuan\inst{1}\textsuperscript{\Letter} \and
Jiali Cui\inst{2} \and
Hanao Li\inst{2}
\and
Tian Han\inst{2}}

\authorrunning{S. Yuan et al.}

\institute{School of Systems and Enterprises,
\and
Department of Computer Science,\\ Stevens Institute of Technology, Hoboken NJ 07310, USA,
\email{\{syuan14,jcui7,hli136,than6\}@stevens.edu}\\
}

\maketitle

\begin{abstract}
Multimodal generative models have recently gained significant attention for their ability to learn representations across various modalities, enhancing joint and cross-generation coherence. However, most existing works use standard Gaussian or Laplacian distributions as priors, which may struggle to capture the diverse information inherent in multiple data types due to their unimodal and less informative nature. Energy-based models (EBMs), known for their expressiveness and flexibility across various tasks, have yet to be thoroughly explored in the context of multimodal generative models. In this paper, we propose a novel framework that integrates the multimodal latent generative model with the EBM. Both models can be trained jointly through a variational scheme. This approach results in a more expressive and informative prior, better-capturing of information across multiple modalities. Our experiments validate the proposed model, demonstrating its superior generation coherence.\\

\end{abstract}   
\section{Introduction}
Generative model (GM) has made remarkable progress in generating high-quality image \cite{ho2020denoising,song2020score}, text \cite{lin2017adversarial,guo2018long}, and video \cite{gupta2023photorealistic}, and recently, the multimodal GM \cite{baltruvsaitis2018multimodal,li2019controllable,zhang2021cross,lin2021vx2text,saharia2022photorealistic,nichol2021glide,gu2022vector} has garnered significant interest for providing a powerful framework that intrigues popular applications of \textit{cross generation}, such as the image-to-text \cite{li2021align}, text-to-image \cite{radford2021learning, nichol2021glide}. However, these prior advances often focus on cross-modality learning by modeling the conditional dependency from one to the other while ignoring learning meaningful semantic representations shared across multimodalities. Learning a shared representation can play a critical role in enabling the downstream tasks \cite{jiang2023understanding}, such as \textit{joint generation}, thus representing an active ongoing research area. 

To tackle the challenge of learning data representations, various methods of latent generative models \cite{kingma2013auto,vahdat2020nvae,NIPS2016_6ae07dcb} have been explored. In particular, the latent variable generative model consists of low-dimensional latent variables and a generation network, where the latent variables are learned to capture the data representation, and thus, the generation network can construct the high-dimensional data by these learned latent variables. For multimodality learning, the multimodal VAEs \cite{wu2018multimodal, shi2019variational},  have recently been developed. Given a set of modalities, these models primarily focus on encoding (inferring) latent variables from different modalities and fusing them into a single latent space. Specifically, MVAE \cite{wu2018multimodal} and MMVAE \cite{shi2019variational} factorize \textit{joint posterior} that infers independent latent variables from different modalities and fuses them into one latent space. Such a single latent space thus needs to represent the data representation of multiple modalities and aims to capture their shared representations. However, these multimodal VAEs only consider less informative Gaussian or Laplacian prior to modeling the latent distribution, which can be limited in expressivity for complex data representations \cite{pang2020learning}, resulting in an ineffectively learned latent generative model.

The energy-based model (EBM), on the other hand, is shown to be expressive and known to be powerful in modelling complex data distribution \cite{gao2020learning, du2020improved}. In high-dimensional data space, it can be difficult for EBM learning as it typically involves Markov Chain Monte Carlo (MCMC) sampling for the EBM density \cite{yuan2024learning, cui2023nips, han2020joint}, while for latent space, the EBM can be formulated as the EBM prior \cite{pang2020learning,XiaoH22}, reducing the burden of EBM prior sampling. Specifically, the EBM prior can be represented as an exponential tilting of the less informative reference distribution, where the energy function serves as a correction of reference distribution (e.g., Gaussian or Laplacian), rendering a more expressive prior distribution. However, most existing EBM works only consider modelling a single modality, and for multimodality, the latent generative model with the EBM prior still remains under-developed. 

In this paper, we intend to explore the EBM prior to the more challenging multimodal learning task. In particular, we present a joint framework that is capable of leveraging the expressivity of EBM prior for modelling the shared semantic latent space from different multimodalities. With a set of modalities, we factorize a latent generative model with the EBM prior, in which the latent representation extracted from different modalities can be well-captured by the EBM prior. Compared to the uni-modal Gaussian or Laplacian prior, the EBM prior can be multi-modal and render more modelling capacity for complex representation learning, which in turn improves the generative power of the whole model and thus maintains semantic coherence for generated samples. Learning such EBM prior typically requires samples from the EBM prior and generator posterior, which is usually achieved by performing MCMC sampling for the EBM prior and generator posterior distribution. However, for the multimodal learning task, MCMC posterior sampling can be time-consuming as it may involve an additional inner-loop for computing the gradient through multiple generation networks from different modalities.

To ensure efficient posterior sampling and facilitate EBM prior learning, we employ the variational learning scheme and introduce an inference model \cite{kingma2013auto,shi2019variational,wu2018multimodal} to approximate the generator posterior. With the inference model matching with the generator posterior, we can directly sample from the inference model and thus circumvent the burden of conducting MCMC posterior sampling. The EBM prior sampling can be achieved by MCMC sampling because of the low-dimensional latent space and, more importantly, the lightweight energy function used. We demonstrate the proposed method can render superior performance in various benchmarks.

Our contributions can be summarized as follows:
\begin{itemize}
  \item We propose the energy-based prior model for multimodal latent generative models to capture complex shared information within multiple modalities.
  \item We develop the variational training scheme where the generation model, inference model, and energy-based prior can be jointly and effectively learned. 
  \item We conduct various experiments and ablation studies and demonstrate superior performance compared to Laplacian prior baselines.
\end{itemize}
\section{Related Works}
\label{sec:related}
In this section, we present the background of multimodal variational autoencoders and energy-based models in detail.\\

\noindent \textbf{Multimodal Variational Autoencoders:} Multimodal Variational Autoencoders (VAEs) \cite{wu2018multimodal,shi2019variational} are based upon the standard VAE and have become the building blocks of all later works. Specifically, in these works, multiple pairs of encoder-decoder for the multi-modalities are learned, with the joint posterior obtained through the particular design of \textit{product-of-experts} or \textit{mixture-of-experts}. Inspired by such idea, mmJSD\cite{sutter2020multimodal}, MoPoE\cite{sutter2021generalized}, MVTCAE\cite{hwang2021multi}, and MMVAE$+$ \cite{palumbo2023mmvae+} have explored VAE-based methods further to tackle the multimodal learning problem. mmJSD \cite{sutter2020multimodal} adopted dynamic prior and modality-specific latent subsets, MVTCAE \cite{hwang2021multi} used multi-view correlation shared representation built upon product of expert \cite{wu2018multimodal}, MoPoE \cite{sutter2021generalized} includes subsets based on both product and mixture of expert \cite{wu2018multimodal, shi2019variational}, and recent baseline MMVAE$+$ \cite{palumbo2023mmvae+} incorporates modality-specific priors that are based on the mixture of expert\cite{shi2019variational}. However, these previous works used less informative uni-modal prior (e.g., Gaussian or Laplacian), which can be ineffective in capturing the complex latent representation shared across different modalities. To tackle this challenge, we propose learning latent space energy-based prior model, which can be more expressive in capturing the shared latent representation and rendering better synthesis across multi-modalities.\\

\noindent \textbf{Energy-Based Models:} The energy-based model (EBM) \cite{xiao2020vaebm,gao2020learning} offers a flexible framework for approximating complex data distributions and is shown to be expressive in capturing the data regularities. In addition to data space EBMs, \cite{pang2020learning} proposes to apply EBM on latent space, which is capable of improving the generative power of the whole model. Learning such latent space EBM requires MCMC posterior and prior sampling, for which the MCMC posterior sampling can be computationally expensive as it requires inner loops of computing the backward gradient of the generation model. Compared to LEBM \cite{pang2020learning} that only studies one (single-) modality, our work targets the challenging multimodal learning problem by learning latent space EBM to capture the shared latent representations. For multimodal learning problems, MCMC posterior sampling can be more difficult as it involves multiple generation networks during the learning. To alleviate the burden of MCMC posterior sampling, we further develop a variational learning scheme to facilitate efficient EBM sampling and learning. We show that the expressive EBM prior can be useful in the challenging multi-modal learning task.
\section{Methodology}
\label{sec:method}
In this paper, we present a novel framework for effectively and efficiently modelling the multimodalities. In particular, we study learning the expressive EBM prior to capturing the shared and complex information across different modalities, for which the less informative Gaussian or Laplacian prior model can be limited in expressivity to effectively model. To facilitate efficiency for our EBM learning and sampling, we further develop a variational learning scheme that incorporates both the generator and inference model for jointly learning with the proposed EBM prior.

\subsection{Energy-based Prior for Multimodalities}
Let $\rvz$ be the latent variable and $\rvX$ be the observation example that contains $m$ modalities, i.e., $\rvX=\{\rvx^{(1)}, \dots, \rvx^{(m)}\}$. A joint distribution can be specified as
\begin{equation}
\label{joint}
\begin{aligned} 
    &p_{\beta, \alpha}(\rvX, \rvz) = p_{\beta}(\rvX|\rvz)p_{\alpha}(\rvz) \;\;\; \text{where}\\
    p_{\beta}(\rvX|\rvz) &= p_{\beta_{(1)}}(\rvx^{(1)}|\rvz)p_{\beta_{(2)}}(\rvx^{(2)}|\rvz)\cdots p_{\beta_{(m)}}(\rvx^{(m)}|\rvz)
\end{aligned}
\end{equation}
in which $p_{\beta}(\rvX|\rvz)$ ($\beta$ collect $\{\beta_{(1)},\dots,\beta_{(m)}\}$) and $p_{\alpha}(\rvz)$ are the generation model and  prior model parameterized by $\beta$ and $\alpha$, respectively. This factorization considers $m$ modalities to be conditionally independently distributed while sharing the same latent space. The modality-common information is modelled by shared latent space $p_\alpha(\rvz)$, and the modality-specific information is modelled by each $p_{\beta_{(i)}}(\rvx^{(i)}|\rvz)$.\\

\noindent\textbf{Generation model.} The $p_{\beta}(\rvX|\rvz)(=\prod_{i=1}^m p_{\beta_{(i)}}(\rvx^{(i)}|\rvz))$ consists of multiple generation models that seek to explain the high-dimensional $\rvx^{(i)}\in \mathbb{R}^{D^{(i)}}$ by a shared low-dimensional latent vector $\rvz \in \mathbb{R}^{d}$ ($d < D^{(i)}$), i.e.,
\begin{equation}
\label{generation-model}
    \rvx^{(i)} = G_{\beta_{(i)}}(\rvz) + \epsilon \;\;\;\; \text{where} \;\;\;\; \epsilon \sim \N(0, I_{D^{(i)}})
\end{equation}
which implies $p_{\beta_{(i)}}(\rvx^{(i)}|\rvz) \sim \N(G_{\beta_{(i)}}(\rvz), I_{D^{(i)}})$ with $G_{\beta_{(i)}}$ being a top-down neural network that maps from $\rvz$ to $\rvx^{(i)}$. We adopt such a generation model for its simplicity (also adopted in \cite{kingma2013auto,pang2020learning,XiaoH22}), but it can also be other choices \cite{NIPS2016_6ae07dcb}.

With a set of $m$ multimodalities $\rvX=\{\rvx^{(1)}, \dots, \rvx^{(m)}\}$, each generation model is designed to be modal-specific, forming the joint distribution $p_{\beta_{(i)},\alpha}(\rvx^{(i)},\rvz)=p_{\beta_{(i)}}(\rvx^{(i)},\rvz)p_{\alpha}(\rvz)$. If $\rvz \sim p_{\alpha}(\rvz)$ can successfully capture the shared information from different modalities, $\rvx^{(i)} \sim p_{\beta_{(i)},\alpha}(\rvx^{(i)},\rvz)$ and $\rvx^{(j)} \sim p_{\beta_{(j)},\alpha}(\rvx^{(j)},\rvz)$ can be drawn with semantic coherence.\\

\noindent\textbf{Energy-based prior.} To effectively capture the semantic information shared across the multimodalities, we intend to learn an expressive prior model. In particular, we study learning the energy-based prior model defined as
\begin{equation}
\label{prior}
    p_{\alpha}(\rvz) = \frac{1}{\mathbb{Z}(\alpha)}\exp [f_\alpha(\rvz)] p_0(\rvz)
\end{equation}
where $\mathbb{Z}(\alpha)$ ($=\int_\rvz \exp[f_\alpha(\rvz)]p_0(\rvz)d\rvz$) is the normalizing constant or partition function, $f_\alpha(\cdot)$ is the energy function parameterized with $\alpha$, and $p_0(\rvz)$ is the referenced distribution usually assumed to be standard Laplacian \cite{shi2019variational, sutter2020multimodal, palumbo2023mmvae+}. Such EBM prior generative model has seen success in modelling the data distribution of \textit{single} modality \cite{pang2020learning,XiaoH22} while in this paper, we intend to explore its effectiveness in modelling the multimodalities.

For multimodalities, capturing the shared content across modalities may serve as the key ingredient toward generating semantic coherent samples. \cite{wu2018multimodal, shi2019variational, sutter2020multimodal, sutter2021generalized} adopt different frameworks for multimodal learning but only consider the less informative, uni-modal Laplacian prior, which in turn limits the model expressivity, leading to an ineffectively learned model. The proposed latent generative model, on the other hand, is learned with the EBM prior, which is known to be powerful in capturing the data regularity and complex distribution. With well-captured shared information for multimodalities, generated samples of each modality thus can maintain strong semantic coherence across different modalities.

\subsection{Learning and Sampling}
\textbf{Maximum likelihood estimation.} Given $n$ observed examples $\{\rvX_1, \dots, \rvX_n\}$ with each $\rvX_i$ containing $m$ modalities $\rvX_i = \{\rvx^{(1)}_i, \dots, \rvx^{(m)}_i\}$, the generator model (Eqn. \ref{joint}) can be learned by maximum likelihood estimation (MLE) as
\begin{equation}
\begin{aligned}
\label{logl}
 L(\theta) &= \sum_{i=1}^n \log p_{\theta}(\rvX_i) = \sum_{i=1}^n \log  \int_\rvz p_{\beta}(\rvX_i|\rvz)p_{\alpha}(\rvz)d\rvz  \\
&= \sum_{i=1}^n \log  \int_\rvz p_{\beta_{(1)}}(\rvx_i^{(1)}|\rvz)\cdots p_{\beta_{(m)}}(\rvx_i^{(m)}|\rvz)p_{\alpha}(\rvz)d\rvz
\end{aligned}
\end{equation}
where $\theta$ collect learning parameters $(\beta, \alpha)$. With a large number of $n$, maximizing Eqn. \ref{logl} is equivalent to minimizing the KL-divergence, i.e., $\KL(\p(\rvX)||p_{\theta}(\rvX))$, and learning $\theta$ can be done by computing the gradient as
\begin{equation}
\begin{aligned}
\label{e5}
\frac{\partial}{\partial \theta} L(\theta) = \E_{p_{\theta}(\rvz|\rvX)}[\frac{\partial}{\partial \theta} \log p_{\theta}(\rvX,\rvz)]
\end{aligned}
\end{equation}
which requires samples from the generator posterior \cite{han2017alternating}. To obtain the posterior samples, it can be typically achieved by performing MCMC sampling for $p_{\theta}(\rvz|\rvX) \propto p_{\beta}(\rvX|\rvz)p_\alpha(\rvz)$. However, for the proposed model on the multimodalities task, $p_{\beta}(\rvX|\rvz)$ consists of $m$ of generation models of each modality, which makes the MCMC posterior sampling inefficient.\\

\noindent\textbf{Variational learning scheme.} To ensure efficient learning and posterior sampling, we introduce the inference model $q_\phi(\rvz|\rvX)$ as an approximation model for the generator posterior. To facilitate learning with the generation model for multimodalities, our inference model is defined to be
\begin{equation}
\label{fuse}
q_{\phi}(\rvz|\rvX) = \frac{1}{m}\sum_{i=1}^m  q_{\phi_{(i)}}(\rvz|\rvx^{(i)})
\end{equation}
where $q_{\phi_{(i)}}(\rvz|\rvx^{(i)})\sim \N(u_{\phi_{(i)}}(\rvx^{(i)}), V_{\phi_{(i)}}(\rvx^{(i)}))$. Such an inference model serves as the \textit{mixture of experts} and is also adopted in \cite{shi2019variational}. Specifically, \cite{shi2019variational} parameterize pairs of generation and inference model to be modal-specific, i.e., $\beta=(\beta_1, \dots, \beta_m)$ and $\phi=(\phi_1, \dots, \phi_m)$, such that each $p_{\beta_i}(\rvx^{(i)}|\rvz)$ and $q_{\phi_i}(\rvz|\rvx^{(i)})$ only focus on one modality as a pair of \textit{decoder} and \textit{encoder}. We adopt such parameterization for its effectiveness. 

With such introduced inference model $q_{\phi}(\rvz|\rvX)$, a joint KL-divergence can be minimized,
\begin{equation}
\label{obj_new}
-L(\beta,\phi,\alpha) = \KL(\p(\rvX)q_{\phi}(\rvz|\rvX)||p_{\beta}(\rvX|\rvz)p_{\alpha}(\rvz))
\end{equation}
which is equivalent to maximizing
\begin{equation}
\label{obj_new_max}
L(\beta,\phi,\alpha) = \frac{1}{m}\sum_{i=1}^{m}\E_{q_{\phi_{(i)}}(\rvz|\rvx^{(i)})}[\log \frac{p_{\beta,\alpha}(\rvX,\rvz)}{q_{\phi}(\rvz|\rvX)}]
\end{equation}
Therefore, we compute the gradient $\frac{\partial}{\partial \beta,\phi}L(\beta,\phi,\alpha)$ as,
\begin{eqnarray}
\label{learn_inf}
\frac{\partial}{\partial \beta,\phi}\frac{1}{m}\sum_{i=1}^{m} [&&\E_{q_{\phi_{(i)}}(\rvz|\rvx^{(i)})}[\log p_{\beta_{(i)}}(\rvx^{(i)}|\rvz)]+\sum_{\substack{j=1,j\neq i}}^{m} \E_{q_{\phi_{(i)}}(\rvz|\rvx^{(i)})}[\log p_{\beta_{(j)}}(\rvx^{(j)}|\rvz)]\nonumber\\
&&+\E_{q_{\phi_{(i)}}(\rvz|\rvx^{(i)})}[\log \frac{p_{\alpha}(\rvz)}{q_{\phi}(\rvz|\rvX)}]]
\end{eqnarray}
the gradient $\frac{\partial}{\partial \alpha}L(\beta,\phi,\alpha)$ is computed as
\begin{equation}\label{learn_ebm}
\frac{\partial}{\partial \alpha}L(\beta,\phi,\alpha)=\E_{q_{\phi}(\rvz|\rvX)}\left[\frac{\partial}{\partial \alpha} f_{\alpha}(\rvz)\right] - \E_{p_\alpha(\rvz)}\left[\frac{\partial}{\partial \alpha} f_{\alpha}(\rvz)\right]
\end{equation}
where $\rvz \sim q_{\phi}(\rvz|\rvX)$ is inferred from the inference model (Eqn. \ref{fuse}), which is a fused joint posterior (i.e., $\rvz \sim \frac{1}{m}\sum_{i=1}^m  q_{\phi_{(i)}}(\rvz|\rvx^{(i)})$ as average weighted over inferred latent vectors from all modalities).\\

\noindent\textbf{Sampling from EBM prior.} Learning EBM (Eqn. \ref{learn_ebm}) requires samples from the EBM prior, which can be accomplished by conducting MCMC sampling, such as Langevin dynamics (LD) \cite{neal2011mcmc}. It iterates as 
\begin{equation}
\label{LD}
     \rvz_{\tau+1} = \rvz_{\tau} + \frac{s^2}{2}\frac{\partial}{\partial \rvz}[\log p_\alpha(\rvz_{\tau})] + s\cdot\epsilon_{\tau}\;\;\; \text{where}\;\;\;  \epsilon_{\tau}\sim\N(0,I_d)
\end{equation}
where $s$ is the step size, $\epsilon$ is the Gaussian noise, and $\tau$ is the time step of Langevin dynamics. As $s\rightarrow 0$, and $\tau \rightarrow \infty$, the marginal distribution of $\rvz$ can asymptotically converge to the target $p_\alpha(\rvz)$ as the stationary distribution. In this work, we employ Laplacian as the initial distribution (i.e. $\rvz_0 \sim \N(0, I_d)$) and conduct short-run Langevin dynamics, which can also provide meaningful learning signals \cite{nijkamp2019learning,nijkamp2020anatomy}.\\

\noindent \textbf{Connection to ELBO.} The VAEs compute the evidence lower bound (ELBO) as the learning objective, which is
\begin{equation}
\label{elbo_vae}
\begin{aligned}
   \E_{q_{\phi}(\rvz|\rvX)}\left[\log p_{\beta}(\rvX|\rvz)\right] - \KL(q_{\phi}(\rvz|\rvX)||p_{0}(\rvz))
\end{aligned}
\end{equation}
where $p_{0}(\rvz)$ is usually assumed to be Gaussian. Whereas, our objective $L(\beta,\phi,\alpha)$ (Eqn. \ref{obj_new}) can be decomposed into the form
\begin{eqnarray}
\label{eblp_ebm}
   L(\beta,\phi,\alpha) = \E_{q_{\phi}(\rvz|\rvX)}\left[\log p_{\beta}(\rvX|\rvz)\right] - \KL(q_{\phi}(\rvz|\rvX)||p_{\alpha}(\rvz))
\end{eqnarray}
which is closely related to the ELBO of VAEs.

Different from Gaussian or Laplacian prior $p_{0}(\rvz)$ in Eqn. \ref{elbo_vae}, we consider learning the EBM prior $p_{\alpha}(\rvz)$. With a set of $m$ modalities, each inference model $q_{\phi_{(i))}}(\rvz|\rvx^{(i)})$ together forms a \textit{mixture of experts}, and thus the Gaussian or Laplacian prior can be limited in expressivity to capture and match with the posterior distribution, while the EBM prior can be more expressive and multi-modal, leading to a well-learned shared latent space. 

\begin{algorithm}[ht]
   \caption{Learning Energy-Based Multimodal Model}
   \label{alg}
\begin{algorithmic}
   \STATE {\bfseries Input:} observation examples $\{\rvx_i^{(1)}, \dots, \rvx_i^{(m)}\}_{i=1}^n$, iteration number $T$, Langevin steps $K$ and step size $s$. iteration step $t=0$
   \REPEAT
   \STATE \textbf{Posterior Sampling}: Given $\{\rvx_i^{(1)}, \dots, \rvx_i^{(m)}\}$, obtain $\{\rvz_i^{(1)}, \dots, \rvz_i^{(m)}\}$ from inference model (Eqn. \ref{fuse}).
   \STATE \textbf{Prior Sampling}: Obtain EBM prior samples $\rvz^{-}$ by performing Langevin dynamics (Eqn. \ref{LD}) with $K$ and $s$. 
   \STATE \textbf{Learn $\phi$ and $\beta$}: Update learning parameter $\phi$, $\beta$ with $\{\rvx_i^{(1)}, \dots, \rvx_i^{(m)}\}$ and $\{\rvz_i^{(1)}, \dots, \rvz_i^{(m)}\}$ using Eqn. \ref{learn_inf}.
    \STATE \textbf{Learn $\alpha$}: Update learning parameter $\alpha$ with inferred samples $\{\rvz_i^{(1)}, \dots, \rvz_i^{(m)}\}$ using Eqn. \ref{fuse} and $\rvz^{-}$ using Eqn. \ref{learn_ebm}.
   \STATE Let $t = t + 1$;
   \UNTIL{$t = T$}
\end{algorithmic}
\end{algorithm}

\section{Experiments}
\label{sec:exp}
In this section, we conduct various experiments to demonstrate the expressiveness of the EBM prior in capturing shared latent information across multimodalities more effectively than less-informative unimodal priors (e.g., Gaussian or Laplacian priors). In our experiments, we follow the prior arts \cite{pang2020learning} and train our model on standard multimodal datasets, such as the PolyMNIST \cite{sutter2021generalized} and MNIST-SVHN \cite{shi2019variational}. We evaluate the performance of our model in terms of joint coherence (\cref{sec:joint-coherence}) and cross coherence (\cref{sec:cross-coherence}). Additionally, we demonstrate the applicability and flexibility of the proposed method, showing its capacity for generalization to various factorizations (\cref{sec:model-generation}), including the incorporation of model-specific factors \cite{palumbo2023mmvae+}. And we visualize the Markov transition on image synthesis and corresponding
generated text applied to Caltech UCSD Birds (CUB) dataset \cite{wah2011caltech}. Furthermore, we conduct ablation studies (\cref{sec:ablation-studies}) to gain a deeper understanding of our approach. We provided our code in  
\url{https://github.com/syyuan2021/Learning-Multimodal-Latent-Generative-Models-with-EBM}\\

\noindent\textbf{Baseline Method.} For comparisons, our direct baselines include MVAE\cite{wu2018multimodal} and MMVAE \cite{shi2019variational}, and we also compare with recent mmJSD \cite{sutter2020multimodal}, MoPoE \cite{sutter2021generalized}, MVTACE\cite{hwang2021multi} and MMVAE$+$ \cite{palumbo2023mmvae+} that are developed based on the foundation methods of \textit{product-of-expert} \cite{wu2018multimodal} and \textit{mixture-of-expert} \cite{shi2019variational}. For a fair comparison, we adopt the same generation and inference network structures, as well as the pre-trained classifiers from \cite{sutter2021generalized} for PolyMNIST, and follow \cite{shi2019variational}'s implementation to train the classifier for MNIST-SVHN.\\

\subsection{Joint Coherence}\label{sec:joint-coherence}
With a set of $m$ modalities, we asses our model in generating \textit{unconditional} image synthesis that maintains strong coherence across different modalities. If the EBM prior is well-learned, it should be capable of sampling the latent variables that capture semantic representations and thus generating the image synthesis semantically consistent between multimodalities. We measure such coherence by computing the classification accuracy of generated images of each modality with corresponding classifiers. Higher accuracy indicates better consistency between predicted categories of multimodal synthesis.\\

\begin{figure}[h]
\captionsetup{singlelinecheck=off}
\captionsetup[subfigure]{singlelinecheck=on}
    \centering   
    \captionbox{Qualitative Results of Joint Generation on  MNIST-SVHN (top) and PolyMNIST (bottom)) \label{fig:joint-results-base}}[0.9\textwidth]{
            \subcaptionbox{EBM Prior}[0.42\textwidth][l]{%
                   \includegraphics[width=0.197\textwidth]{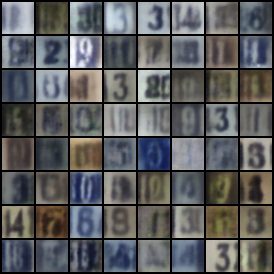}%
                  \includegraphics[width=0.197\textwidth]{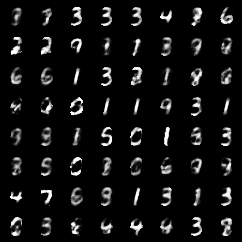}%
               \\    \includegraphics[width=0.394\textwidth]{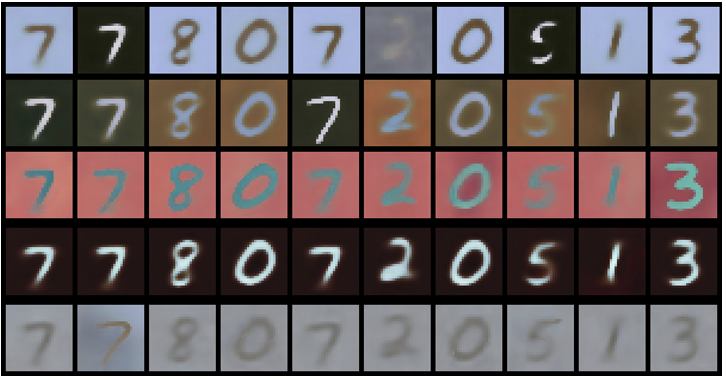}%
            }
            \subcaptionbox{Unimodal Prior: MMVAE$^{*}$}[0.42\textwidth][l]{%
               \includegraphics[width=0.199\textwidth]{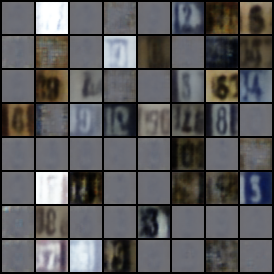}%
              \includegraphics[width=0.199\textwidth]{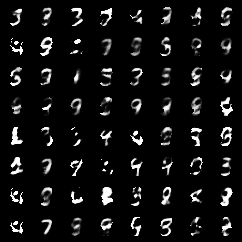}%
               \\    \includegraphics[width=0.398\textwidth]{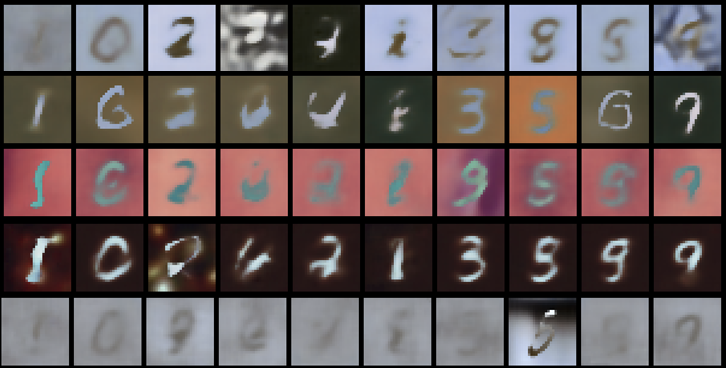}%
            }
        }
\end{figure}

We report the quantitative results in Tab. \ref{tab:gen-base}, where our EBM prior shows superior performance compared to baseline models. Both numerically and visually, the EBM prior successfully captures complex latent representations, whereas the unimodal prior (Laplacian) is less expressive and limited in modeling meaningful semantic representations. Synthesis of joint generation can be found in Fig. \ref{fig:joint-results-base}. \\

\begin{table}[h]
    \centering
    \caption{Accuracy of Joint Coherence and Cross Coherence for PolyMNIST and MNIST-SVHN datasets. The PolyMNIST results for MVAE are referenced from \cite{palumbo2023mmvae+}, and MNIST-SVHN results for MVAE are referenced from \cite{shi2019variational}. MMVAE$^{*}$ refers to \textit{mixture of expert} without important sampling. }
    \label{tab:gen-base}
    \resizebox{0.8\columnwidth}{!}{
    \begin{tabular}{c|c|c|c|c|c}
        \toprule
        \multicolumn{3}{c|}{Joint Coherence} & \multicolumn{3}{c}{Cross Coherence} \\
        \midrule
        \multirow{2}{*}{Model} & \multirow{2}{*}{PolyMNIST}  & \multirow{2}{*}{MNIST-SVHN} & \multirow{2}{*}{PolyMNIST} & \multicolumn{2}{c}{MNIST-SVHN} \\
        \cmidrule(lr){5-6}
        & & & & M->S & S->M \\
        \midrule
        Ours &\textbf{0.746} $\uparrow$ &\textbf{0.419} $\uparrow$ &\textbf{0.853} $\uparrow$ &\textbf{0.237} $\uparrow$ &\textbf{0.653} $\uparrow$\\
        \midrule
        MVAE &0.080 &0.127 &0.298 &0.095 &0.093 \\
        \midrule
        MMVAE$^{*}$ &0.232 &0.215 &0.844 &0.169 &0.523 \\
        \bottomrule
    \end{tabular}
    }
\end{table}

\subsection{Cross Coherence}\label{sec:cross-coherence}
Next, we evaluate our model in cross-modal generation. Cross-modal generation is to generate image synthesis from input modality $\rvx^{(i)}$ to target modality $\rvx^{(j)}$ ($i \ne j$). Specifically, given input $\rvx^{(i)}$, we first obtain inferred latent vector $\rvz \sim q_\phi(\rvz|\rvx^{(i)})$ and then generate synthesis via generation model $\rvx^{(j)} \sim p_{\beta_{j}}(\rvx^{(j)}|\rvz)$. With our EBM prior, the inference and generation model can be learned to integrate both the modality bias and the energy-based refinement and thus the latent vectors inferred from $\rvx^{(i)}$ carry the semantic representation shared with $\rvx^{(j)}$, leading to coherent input $\rvx^{(i)}$ and output $\rvx^{(j)}$. To measure the coherence, we compute the classification accuracy (same classifiers used in Sec. \ref{sec:joint-coherence}) of the predicted category of $\rvx^{(j)}$ and true category of $\rvx^{(i)}$. Higher accuracy means both the $\rvx^{(i)}$ and $\rvx^{(j)}$ share higher similarity in semantic features (e.g., digit classes in PolyMNIST). The results can be found in Tab. \ref{tab:gen-base}, in which the proposed method can render competitive performance compared to the baseline methods, suggesting the effectiveness of our EBM prior in multimodalities learning.

\subsection{Model Generalization}\label{sec:model-generation}
This paper studies learning a novel framework for foundation multimodal latent generative models, such as MVAE and MMVAE, which serve as the building blocks of various recent works. In this section, we highlight the applicable capability of the proposed method toward other prior advances \cite{palumbo2023mmvae+}. These advances usually factorize additional \textit{modal-specific prior} or \textit{modal-subset} to benefit the model complexity in learning the complex multimodalities. In particular, MMVAE$+$ develop their latent generative model as
\begin{equation}
\label{mmvae+}
\begin{aligned} 
    p_{\beta}(\rvX, \rvz,\rvW) = p_{\beta}(\rvX|\rvz,\rvW)p_0(\rvz)p_0(\rvW)
\end{aligned}
\end{equation}
where the prior model $p_0(\rvW)$ is modal-specific (i.e., $\rvW=\{\rvw^{(1)},\dots,\rvw^{(m)}\}$) and is introduced to improve the expressivity of the whole prior model (compared to Eqn. \ref{joint}. However, the shared latent representation is still modelled by $p_0(\rvz)$, which is assumed to be Laplacian and can be less informative.\\

\vspace{-5mm}
\begin{table}[h]
\centering
\setlength{\belowcaptionskip}{-0.5pt}%
\caption{Accuracy of Joint Coherence and Cross Coherence for PolyMNIST. Results of MMVAE$+$, MoPoE, MVTCAE, and mmJSD are reported in \cite{palumbo2023mmvae+}. Ours$+$ refers to generalizing our model with modality-specific priors.}
\label{tab:gen-advance}
\resizebox{0.6\columnwidth}{!}{
\begin{tabular}{c|c|c}
    \toprule
    Model 	& Joint Coherence & Cross Coherence \\
    \midrule
    Ours$+$ &\textbf{0.878 $\uparrow$}  &\textbf{0.897 $\uparrow$} \\
    \midrule
    MMVAE$+$ &0.344 &0.869  \\
    MoPoE &0.141 &0.720  \\
    MVTCAE  &0.003 &0.591 \\
    mmJSD   &0.060 &0.778  \\
    \bottomrule
\end{tabular}
}
\end{table}
\vspace{-5mm}

\begin{figure}[ht]
    \centering
    \begin{subfigure}[]{0.49\textwidth}
    \includegraphics[height=0.15\textheight]{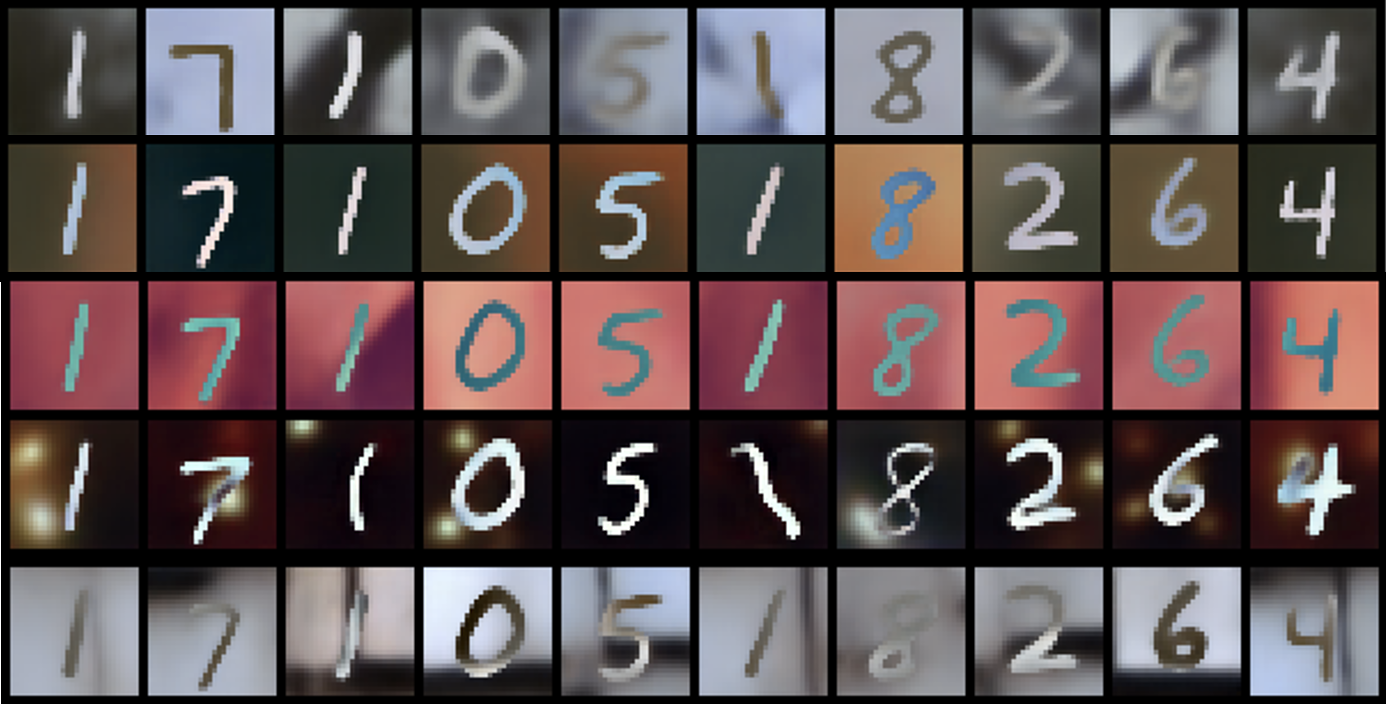}
    \caption{ EBM Prior}
    \end{subfigure}
    \begin{subfigure}[]{0.49\textwidth}
    \includegraphics[height=0.15\textheight]{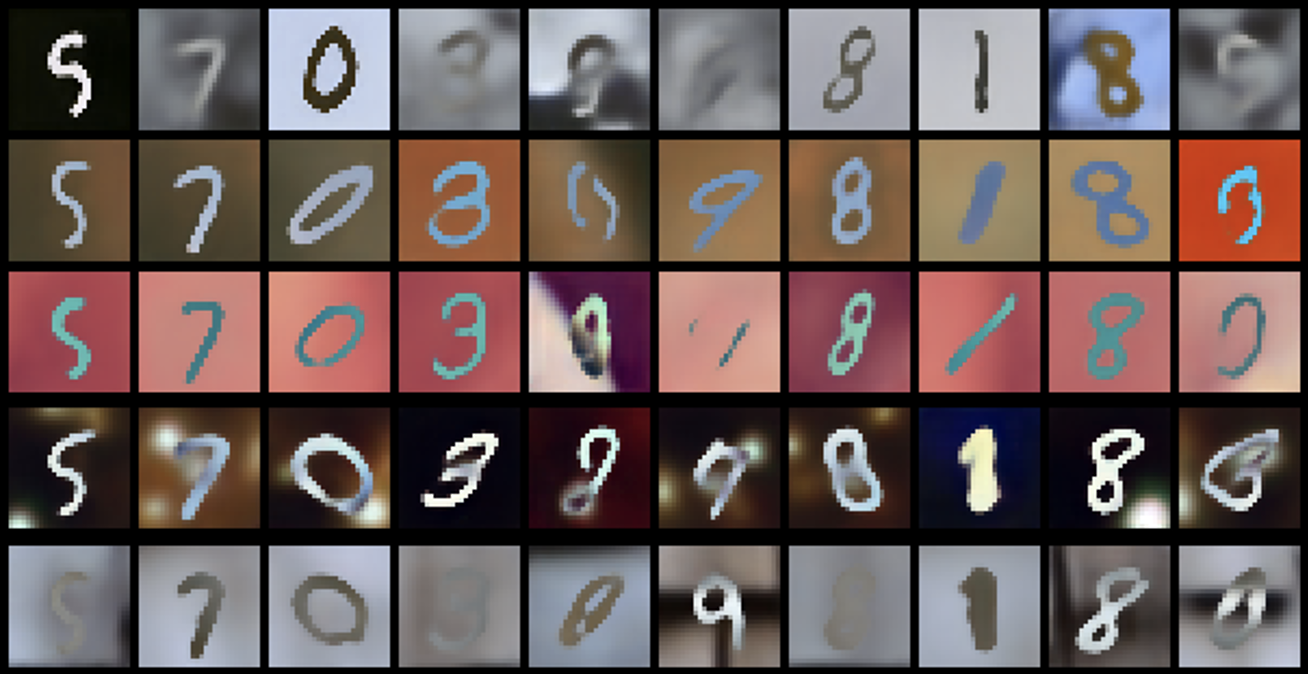}
    \caption{Unimodal Prior}
    \end{subfigure}   
    \caption{Qualitative results of Joint Generation on PolyMNIST}
    \label{fig:joint-results-advance}  
\end{figure}

\begin{figure}[h]
    \centering
    \begin{subfigure}{1\textwidth}
            \centering 
            \includegraphics[width=0.19\textwidth,height=0.4\textheight]{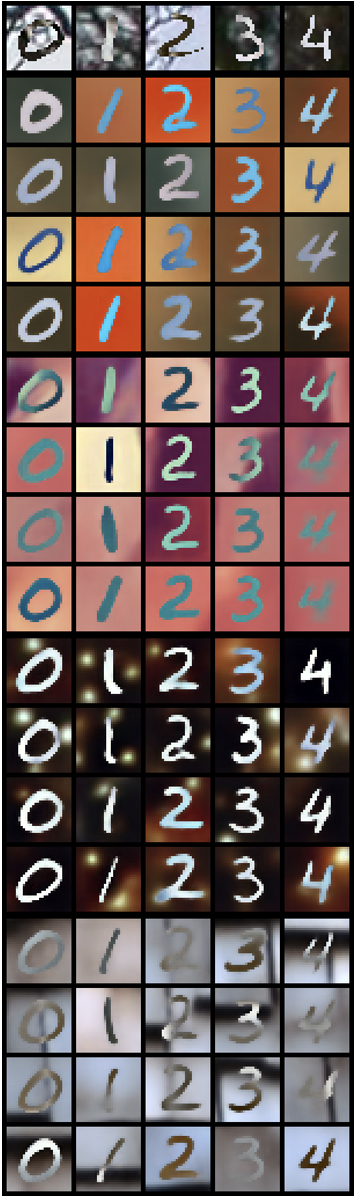}
            \includegraphics[width=0.19\textwidth,height=0.4\textheight]{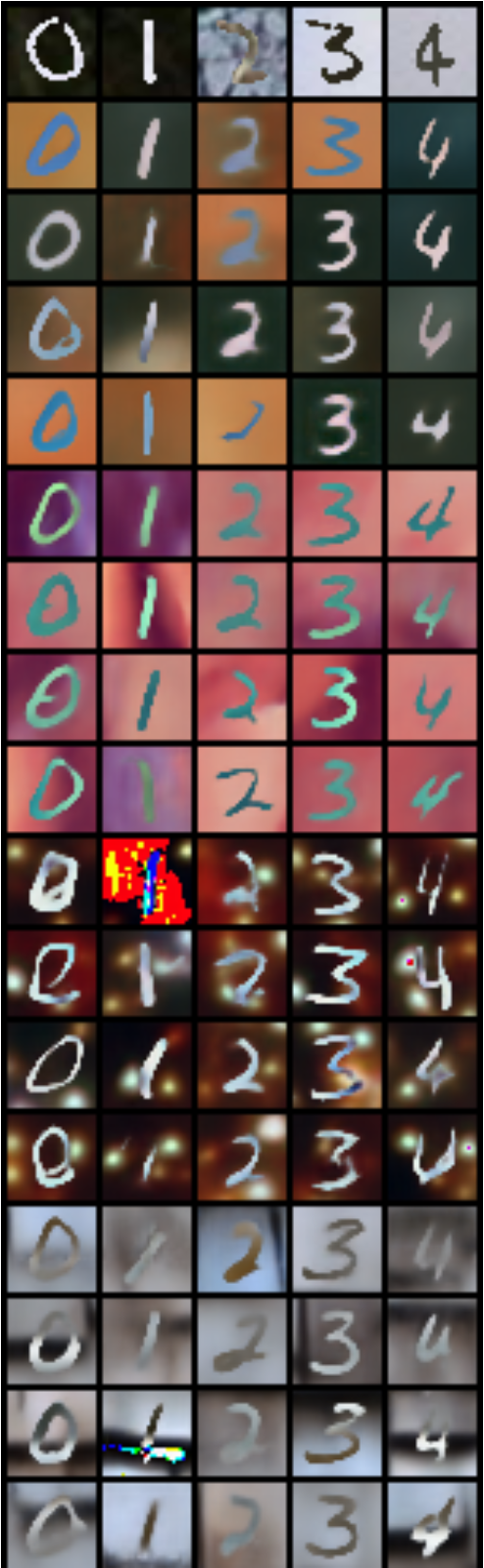}
            \includegraphics[width=0.19\textwidth,height=0.4\textheight]{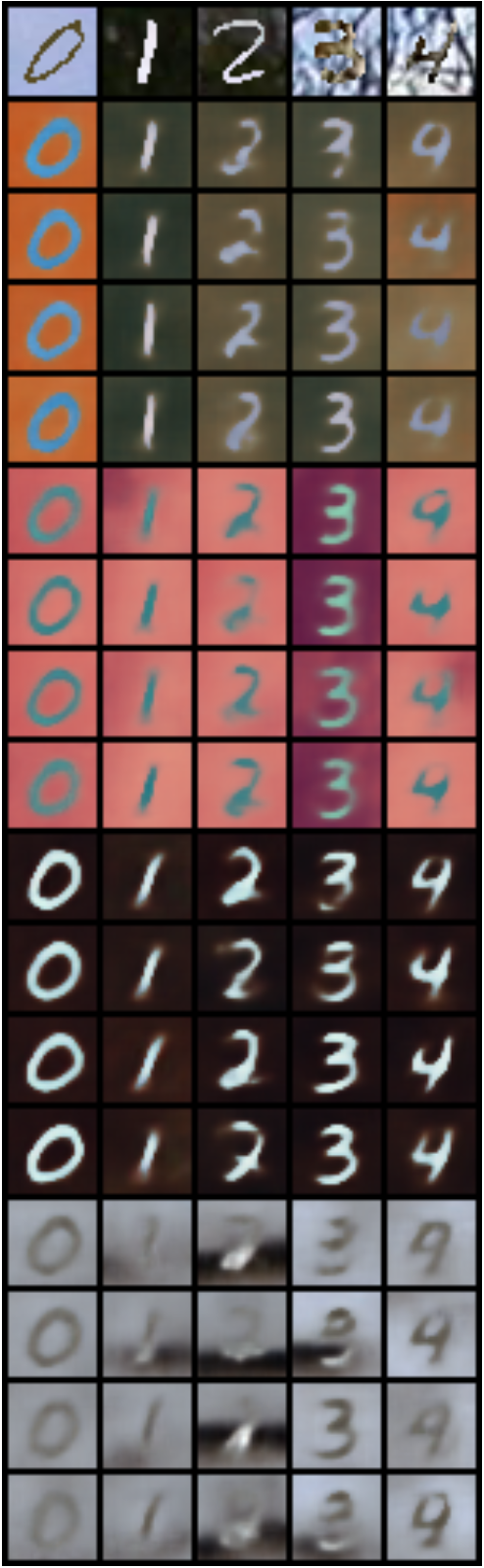}
            \includegraphics[width=0.19\textwidth,height=0.4\textheight]{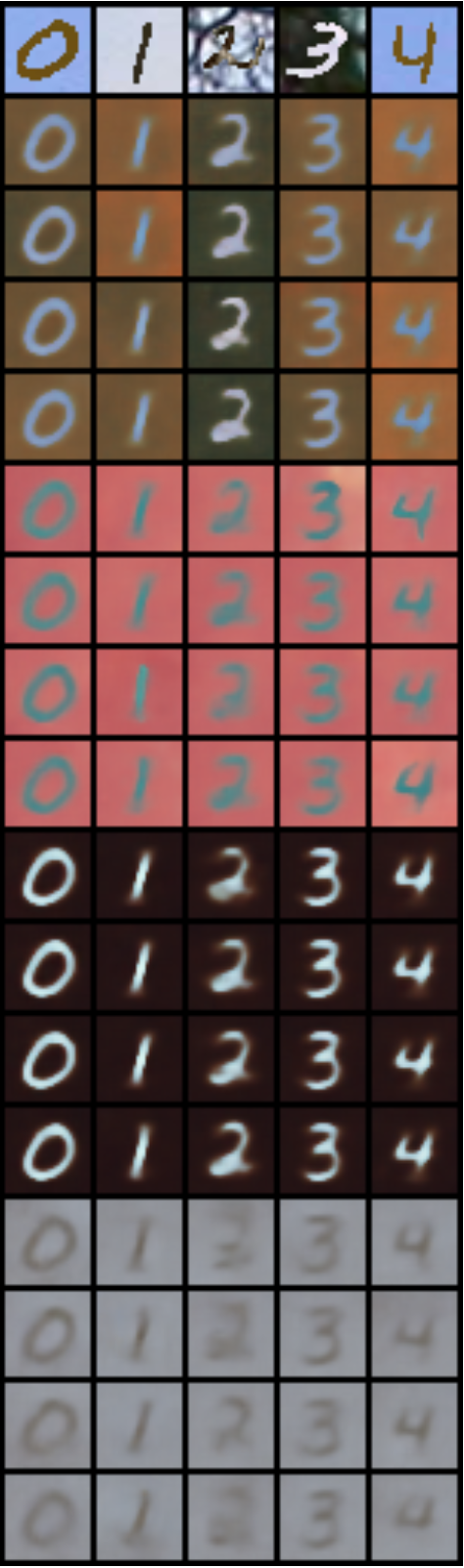}
            \includegraphics[width=0.19\textwidth,height=0.4\textheight]{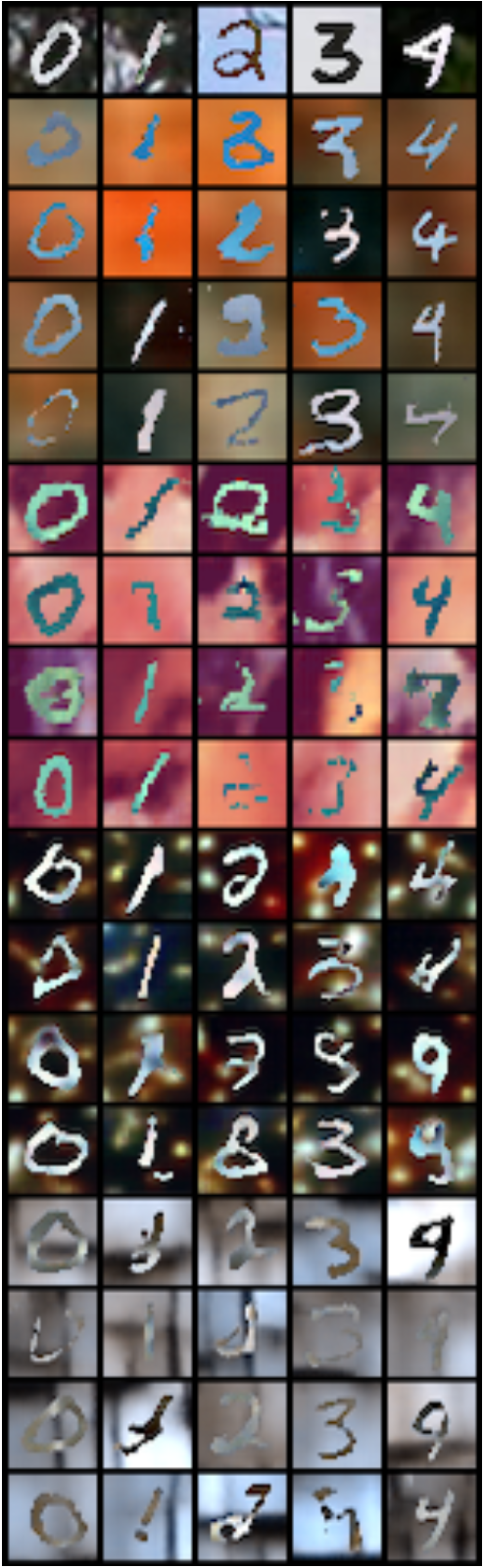}
    \end{subfigure}
    \caption{Qualitative results of Cross Generation on PolyMNIST. From left to right, cross generation results are from (a) \textbf{EBM prior}  (b) MMVAE$+$, (c) MoPoE, (d) mmJSD, and (e) MVTCAE. }
    \label{fig:cross-results-advance}
\end{figure}


\noindent\textbf{Generalization.} Our model is flexible and can be adapted to incorporate such a modal-specific factor, i.e., $p_{\beta,\alpha}(\rvX, \rvz, \rvW) = p_{\beta}(\rvX|\rvz, \rvW)p_{\alpha}(\rvz)p_{0}(\rvW)$, such that the shared latent representation is modelled by the proposed EBM prior $p_{\alpha}(\rvz)$, which can be multi-modal and more expressive than the unimodal prior model $p_0(\rvz)$. To examine the effectiveness, we follow MMVAE$+$ and train our model on PolyMNIST with the same network structures. We report the results of joint coherence and cross coherence in Tab. \ref{tab:gen-advance} and the qualitative results in Fig. \ref{fig:joint-results-advance} and Fig. \ref{fig:cross-results-advance}. \\

\noindent \textbf{Caltech UCSD Birds Dataset (CUB)}. To examine the scalability, we further train our model on the challenging multimodal dataset Caltech UCSD Birds (CUB)\cite{wah2011caltech}. The proposed EBM can be viewed as an exponential tilting of the reference distribution and thus can correct the less informative prior model toward being more expressive. In this section, we intend to demonstrate the expressivity of our EBM prior in correcting the unimodal prior model. In practice, we utilize the same network structure and train MMVAE$+$ on CUB, and then we learn our EBM prior with the pre-trained generation and inference model. If our EBM prior can be learned well, the quality of synthesis generated by our EBM prior samples should be better than samples of the unimodal prior.

\begin{figure}[h]
    \centering
    \includegraphics[width=0.99\textwidth]{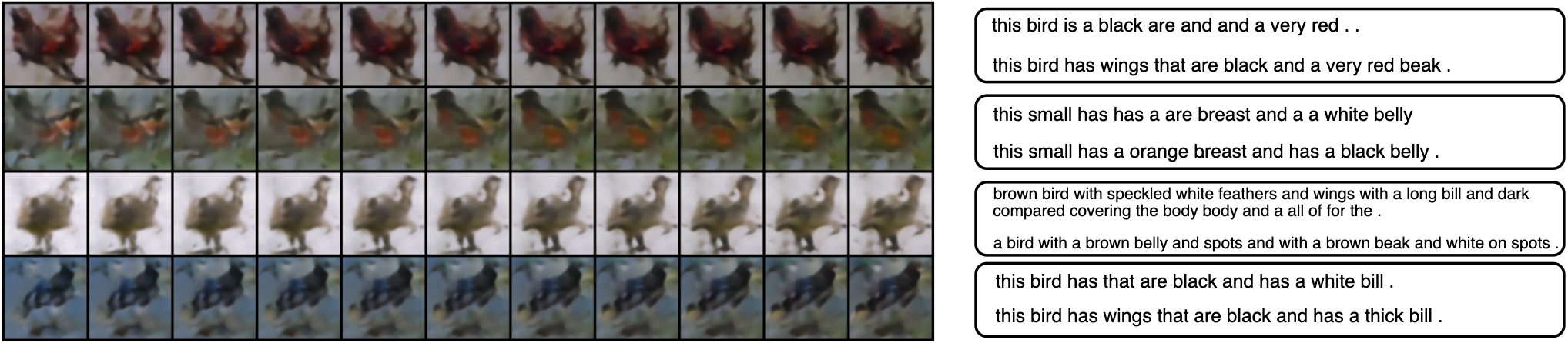}
    \caption{Visualization of Markov chain transition on image synthesis and corresponding generated text. \textbf{Left}: visualization of Markov chain transition where left column images are generated by \textit{unimodal} prior sample, and right column images are generated by \textit{EBM} prior sample. \textbf{Right}: jointly generated text corresponding to each row in left figure. The sentence at the top is generated by  \textit{unimodal} prior sample, and the sentence at the bottom is generated by \textit{EBM} prior sample.}
    \label{fig:cub_mcmc}
\end{figure}

For illustration, we visualize the Markov chain transition during Langevin dynamics on generated images and text in Fig. \ref{fig:cub_mcmc}. Specifically, the Markov chain transition starts from the latent drawn from the unimodal prior $p_0(\rvz)$ and then progresses as an iterative sampler for our EBM prior $p_\alpha(\rvz)$. It can be seen that the quality of image and text synthesis becomes better as the Langevin dynamics progresses and, more importantly, at the final step, the generated images and text render better semantic coherence, which further indicates the effectiveness of the proposed EBM prior.

\vspace{-8mm}
\subsection{Ablation Studies}\label{sec:ablation-studies}
In the previous sections, we demonstrated the effectiveness of using an EBM prior in multimodal contexts, illustrating that our model can better capture complex data representations across different modalities. To further investigate our model's capabilities, we examine the impact of our EBM prior settings on performance through a series of experiments conducted on the PolyMNIST dataset.\\

\noindent \textbf{Energy-Based Model MCMC Steps:} We first examine the influence of the number of MCMC steps, denoted as $S$, when sampling the latent variables from EBM prior. We fixed the network architecture to consist of 4 layers with 400 hidden units each. As observed in the middle two rows of Tab. \ref{tab:abalation_poly}, a smaller number of MCMC steps results in poorer performance. However, increasing the number of MCMC steps is computationally expensive, leading to a trade-off between computation time and model performance. \\

\noindent \textbf{Energy-Based Model Layers:} Then we examine whether more information transformation and interaction will increase coherence scores. So we fixed hidden dimension and Langevin steps while increase EBM layer $L$ from 4 to 6. As shown in the first two rows of Tab. \ref{tab:abalation_poly}, increasing the number of layers $L$ leads to better coherence results. \\

\noindent \textbf{Energy-Based Model Complexity:} 
Increasing the model dimension allows the model to learn more complex and representative features. To examine scalability, we increase the dimension of hidden units $D$ to investigate the performance impact of capacity in capturing and processing information. We maintain the number of layers at 4 and use 50 Langevin steps. As observed in the bottom two rows of Tab. \ref{tab:abalation_poly}, increasing the number of units results in better coherence performance. \\

\vspace{-5mm}
\begin{table}[h]
\centering
\setlength{\belowcaptionskip}{-0.5pt}%
\caption{Ablation studies for energy-based models with different numbers of hidden units $D$, MCMC steps $S$, and network layers $L$.}
\label{tab:abalation_poly}
\resizebox{0.8\columnwidth}{!}{
\begin{tabular}{c|c|c}
    \toprule
    Model 	& Joint Coherence & Cross Coherence \\
    \midrule
    Ours (D=200, L=4, S=50) &0.574 &0.842 \\
    Ours (D=200, L=6, S=50) &0.645 &0.832 \\
    \midrule
    Ours (D=400, L=4, S=30) &0.683 & 0.845\\
    Ours (D=400, L=4, S=50) &\textbf{0.746} &\textbf{0.853} \\
    \midrule
    Ours (D=200, L=4, S=50) &0.574 &0.842 \\
    Ours (D=400, L=4, S=50) &\textbf{0.746} &\textbf{0.853} \\
    \bottomrule
\end{tabular}
}
\end{table}
\vspace{-11mm}

\section{Conclusions and Future Work}
\label{sec:con}
In this paper, we present a novel framework for multimodal latent generative models with an EBM prior. This expressive and flexible prior can better represent multimodal data complexity and capture shared information among modalities. Our experiments demonstrate significantly improved coherence of synthesized samples across different modalities compared to baseline models. Our proposed model also facilitates cross-generation between modalities, as validated by experimental results.

However, the proposed method is based on a simple \textit{mixture of experts} scheme with an EBM prior to optimize the multimodal \text{ELBO}. Approaches that provide a tighter bound on \text{ELBO}, such as importance sampling and stratified sampling with an EBM prior, have not been fully explored. Additionally, to gain a better understanding of shared information learning schemes under multimodal contexts, other expressive priors such as normalizing flow or hierarchical priors will be considered in our future research. For EBM learning, we use latent variables from the variationally inferred posterior, which is less accurate in approximating the true posterior compared to methods such as MCMC sampling. While the latter can be time-consuming, this trade-off either sacrifices generative performance or computational efficiency. This dilemma is non-trivial in multimodal generative problems. Lastly, we will explore the scalability of our EBM prior in future work by investigating its performance on realistic multimodal datasets. This will allow us to evaluate the model's effectiveness in more complex and varied real-world scenarios, further validating its applicability and robustness.


%
%
\bibliographystyle{splncs04}
\bibliography{main}
\end{document}